# A Neural-Network Technique to Learn Concepts from Electroencephalograms

Vitaly Schetinin[1], and Joachim Schult[2]

[1]Compure Science Department, University of Exeter, Exeter, EX4 4QF, UK.
V.Schetinin@ex.ac.uk
[2]Friedrich-Schiller-University of Jena, Jena, Germany
Joachim_Schult@web.de

**Abstract**. A new technique is presented developed to learn multi-class concepts from clinical electroencephalograms. A desired concept is represented as a neuronal computational model consisting of the input, hidden, and output neurons. In this model the hidden neurons learn independently to classify the electroencephalogram segments presented by spectral and statistical features. This technique has been applied to the electroencephalogram data recorded from 65 sleeping healthy newborns in order to learn a brain maturation concept of newborns aged between 35 and 51 weeks. The 39399 and 19670 segments from these data have been used for learning and testing the concept, respectively. As a result, the concept has correctly classified 80.1% of the testing segments or 87.7% of the 65 records.

## 1. Introduction

Machine learning and neural-network techniques [1 - 6] have been successfully used to learn classification models, or concepts, from real-world data including electroencephalograms (EEGs) [7 - 13]. These methods explore the given set of the input variables, or features, assumed to represent the classification problem and discard those which are not relevant to the classification problem, e.g. corrupted by a noise, because such input variables can seriously hurt the generalization ability of the induced concepts [1 – 6, 11 – 13].

To discard the irrelevant features a number of feature selection methods have been suggested some of which are based on a greedy, or hill-climbing, strategy [1, 4, 5]. Being applied to multivariate classification problems the learning methods based on such a strategy are able to find out a sub-optimal set of features for acceptable learning time [4, 5, 11 – 13].

In general the learning time required for learning concepts in the presence of irrelevant features increases proportionally to the data size as well to the number of classes [1 - 6]. The computational time additionally increases if the classification problem is presented by examples which are non-linearly separable and the class boundary overlap each other heavily [1, 5].

On the other hand, the concept induced from the given data, e.g. from the EEGs, has to be observable by users which want to understand a decision making mechanism. For this reason, users may prefer to use the classification models based on decision trees (DTs) which are easy-to-understand and interpret [1 – 6].

The DTs usually consist of the nodes of two types: one is a decision node containing a test, and other is a leaf node assigned to an appropriate class. A branch of the DT represents each possible outcome of a test. An example coming at the root of the DT follows the branches until a leaf node is reached. The name of the class at the leaf is the resulting classification. The node can test one or more of the input features. A DT is multivariate one, if its nodes test more than one of the features. The multivariate DT is much shorter than that which tests a single feature [1 – 6].

To learn concepts presented by numerical attributes the authors [2 - 6] suggested the multivariate DTs with Threshold Logical Units (TLUs) or single neurons. Such multivariate DTs are known also as the oblique DTs.

In this paper we describe a neural-network technique we developed to induce the multi-class concepts from large-scale clinical EEGs. In our research we aim to use the DT classification models to evaluate the brain maturation of newborns whose EEGs have been recorded during sleeping hours. In clinical practice such evaluations of brain maturation can assist clinicians to objectively diagnose some brain pathologies of newborns [10 - 13].

Below in section 2 we describe the classification problem and the clinical EEG data. In section 3 we briefly describe some DT techniques and then in section 4 present a neural-network technique that we developed to induce the multi-class concept from large-scale EEG data. Finally we evaluate the classification accuracy of the induced concept and discuss our results.

## 2. A Classification Problem and the EEG Data

In this section we first describe the classification problem and the structure of EEG data and second present the result of statistical analysis aimed to evaluate the contribution of the input variables to the classification problem.

### 2.1. A Classification Problem

In general, learning multi-class concepts from EEGs is still a difficult problem [7, 9, 10]. First, EEGs are strongly non-stationary signals whose statistical characteristics vary widely; second, the EEGs depend on the background brain activity of patients; third, the EEGs are the weak invoked potentials which are corrupted by noise and mus-cle artifacts and, fourth, user may give some EEG features which are irrelevant to the classification problem.

The EEG data used in our experiments were recorded from 65 sleeping newborns via the standard electrodes C3 and C4. The age of these newborns was between 35 to 51 weeks and, therefore, the EEG data represent 16 age groups, or classes, which differ by one week in average.

Following [11 – 13], we represent each EEG segment by the 72 spectral and statistical features calculated on a 10-sec interval into 6 frequency bands such as sub-delta (0-1.5 Hz), delta (1.5-3.5 Hz), theta (3.5-7.5 Hz), alpha (7.5-13.5 Hz), beta 1 (13.5-19.5 Hz), and beta 2 (19.5-25 Hz). These features were calculated for the standard electrodes C3 and C4 as well as for their sum, C3 + C4, for the real and absolute spectral powers and variances.

Observing these EEG records, EEG-expert has manually removed the muscle artifacts from these records such that after cleaning an average rate of the outlying segments, whose values were more $3\delta$, did not exceed 6%, where $\delta$ is the standard deviation calculated on the patient's record. The total sum of all EEG segments presented by the 72 features is equal to 59069 therefore we can say that such a dataset is large-scale. The statistics of segments over the 16 classes are shown in Table 1 below:

**Table 1:** Statistics of the EEG data

| Class | Age | $N_p$ | $N_s$ |
|---|---|---|---|
| 1 | 35 | 1 | 996 |
| 2 | 37 | 1 | 617 |
| 3 | 38 | 8 | 6379 |
| 4 | 39 | 7 | 037 |
| 5 | 40 | 7 | 4371 |
| 6 | 41 | 5 | 4490 |
| 7 | 42 | 2 | 1113 |
| 8 | 43 | 2 | 3435 |
| 9 | 44 | 8 | 8524 |
| 10 | 45 | 7 | 6884 |
| 11 | 46 | 2 | 1847 |
| 12 | 47 | 5 | 5562 |
| 13 | 48 | 1 | 548 |
| 14 | 49 | 6 | 6446 |
| 15 | 50 | 2 | 1139 |
| 16 | 51 | 1 | 681 |
| Total | | 65 | 59069 |

Here $N_p$ and $N_s$ are the number of patients and segments in the given class.

All 65 newborns were healthy. So for a healthy newborn the concept learnt from these data should assign all the EEG segments to a corresponding age group. But for newborns with some brain development pathologies the output of the classification model expected to be mismatched. Below we analyze some statistical characteristics of these EEG data calculated over all 16 classes.

## 2.2. Statistical Analysis of the EEG Data

To see how the EEG data vary, in Fig 1 below we plot two principal components, $p_1$ and $p_2$, calculated into different time windows for two newborns belonging to different classes. These components were calculated for four subsequent time intervals consisting of 300 segments: 1:300, 301:600, 601:900, and 901:1200.

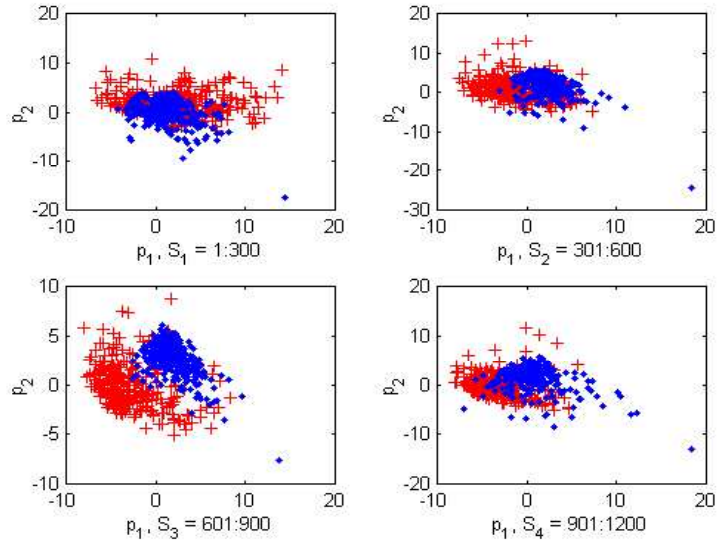

**Fig 1:** EEG segments of two patients into four subsequent intervals

Observing these plots, we can see that the values of $p_1$ and $p_2$ vary over time so that the class boundaries move dramatically. The class boundaries move because the EEGs reflect the individual brain activities of the newborns. This activity has a chaotic character which increases the group variance of the classes. To evaluate the influence of individual activity, we used the following statistical technique.

Let us first introduce a variance $v(x_j)$ of all $r$ classes and a group variance $s_i(x_j)$ of the $i$th class for the input variable $x_j$:

$$v(x_j) = \frac{1}{r}\sum_{k=1}^{r}(\tilde{x}_{kj} - \tilde{y}_j)^2, \ \tilde{y}_j = \frac{1}{r}\sum_{k=1}^{r}\tilde{x}_{kj},$$

$$s_i(x_j) = \frac{1}{N}\sum_{k=1}^{N}(x_{kj} - \tilde{x}_{ij})^2, \ \tilde{x}_{ij} = \frac{1}{N}\sum_{k=1}^{N}x_{kj},$$

where $N$ is the number of the $i$th class segments, $x_{kj}$ is a $j$th feature value of a $k$th example, $\tilde{x}_{ij}$ is a mean value of a $j$th feature calculated for the $i$th class.

Then we can evaluate the significance of the $j$th feature as the next ratio:

$$d_j = 100 \frac{v(x_j)}{\sum_{i=1}^{r} s_i(x_j)}, \quad j = 1, \ldots, m.$$

We see that value of $d_j$ decreases proportionally to the sum of group variances $s_i$ and increases proportionally to the variance $v(x_j)$ of the age groups. Clearly, that group variance $s_i$ grows proportionally to the individual activity of patients belonging to the same group. Therefore we can conclude that the value $d_j$ reflects the significance of feature $x_j$.

Fig. 2 depicts the values of $d$, $v$ and $s$ calculated for all 72 features on the training set. From the top plot in this Fig we can see that $x_{36}$ is the most important and $x_{38}$ is the less impotent features.

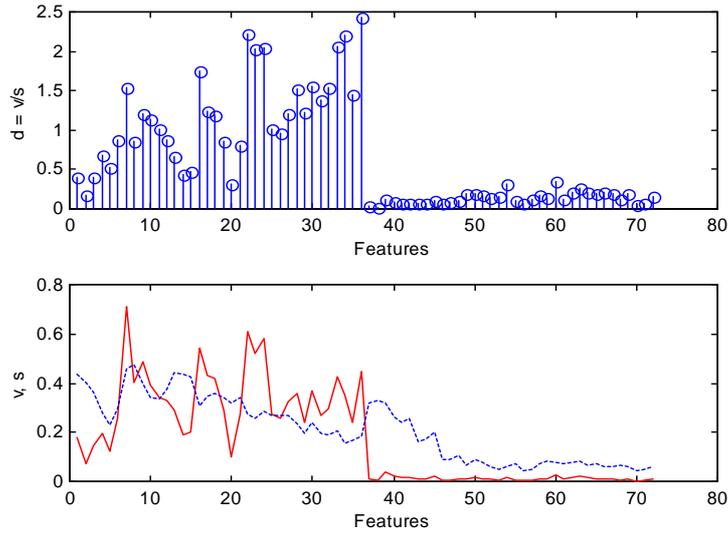

**Fig 2:** The significance of the 72 features

In Fig. 3 we depict the standard $3\sigma$ intervals calculated for the features $x_{36}$ and $x_{38}$. Observing the behavior of these features over 16 classes, we can see that the most important feature $x_{36}$ is dependent on the classes more than feature $x_{38}$. Moreover, the $3\sigma$ interval of the feature $x_{36}$ is less than the corresponding interval of feature $x_{36}$. However, using the feature $x_{36}$ we can not properly distinguish all 16 classes because its $3\sigma$ interval is large yet.

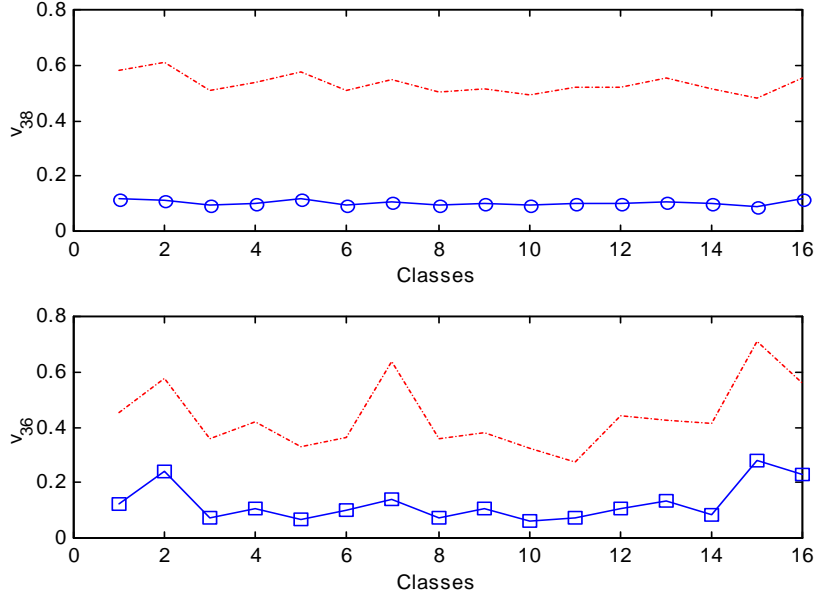

**Fig 3:** The 3σ intervals for features $x_{38}$ and $x_{36}$

Below we describe some decision tree techniques that are applied for such multi-class problems.

## 3. Induction of Decision Trees

In this section we briefly describe a linear machine technique which is applied for multi-class problems and discuss its performance on the EEG data.

### 3.1. A Linear Machine

A Linear Machine (LM) is a set of $r$ linear discriminant functions that are calculated in order to assign an example **x** to one of the $r \geq 2$ classes [1, 5, 6]. Each internal node of the LM tests a linear combination of $m$ input variables $x_1, x_2, \ldots, x_m$.

Let us now introduce an extended input vector $\mathbf{x} = (x_0, x_1, \ldots, x_m)^\mathrm{T}$, where $x_0 \equiv 1$, and a discriminant function $g(\mathbf{x})$. Then the linear test at the $j$th node is described as follows:

$$g_j(\mathbf{x}) = \Sigma_i w_i^j x_i = \mathbf{w}^{(j)\mathrm{T}} \mathbf{x} > 0, \ i = 0, \ldots, m, \ j = 1, \ldots, r,$$

where $w_0^{(j)}, \ldots, w_m^{(j)}$ are the real-valued coefficients of a weight vector $\mathbf{w}^{(j)}$ of $j$th TLU.

The LM assigns an example **x** to the *j*th class if and only if the output of the *j*th node is higher than the outputs of the other nodes:

$$g_j(\mathbf{x}) > g_k(\mathbf{x}), \ j = 1, ..., r, \ k \neq j.$$

Note that this strategy of making decision is known also as a Winner Take All (WTA).

For learning of LMs, the weight vectors $\mathbf{w}^{(j)}$ and $\mathbf{w}^{(k)}$ of discriminant functions $g_j$ and $g_k$ are updated on an example **x** that the LM misclassifies. The weights $\mathbf{w}^{(j)}$, where *j* is the class to which the example **x** actually belongs, should be increased and the weights $w^{(k)}$, where *k* is the class to which the LM erroneously assigns **x**, should be decreased. This is done by using the following error correction rule:

$$\mathbf{w}^{(j)} := \mathbf{w}^{(j)} + c\mathbf{x},$$
$$\mathbf{w}^{(k)} := \mathbf{w}^{(k)} - c\mathbf{x},$$

where $c > 0$ is the amount of correction, or iteration.

If the training examples are linearly separable, the above procedure can train the LM for a finite number of iterations. However, if the training examples are non-linearly separable, this learning rule can not provide predictable classification accuracy. For such cases a Pocket algorithm [1] has been suggested which saves the best result that occurs during learning. Below we describe the application of this algorithm to our EEG problem.

### 3.2. An Application of Linear Machine to the EEG Data

The large-scale EEG data that we aim to classify split into 16 classes. The centers of these classes differ, in average, by one week and lay extremely close each other so that they overlap hardly. These circumstances deteriorate the performance of the Pocket Algorithm significantly and we found out that its classification accuracy not exceed 71.6% of the testing data. We can assume the following reason of such small accuracy.

When the algorithm classifies a training example **x** erroneously, it updates weights of two linear tests $\mathbf{w}^{(j)}$ and $\mathbf{w}^{(k)}$ and sets the length of the correctly classified example sequence to zero. For the next training examples the algorithm will evaluate the accuracy of the LM on all the training examples. So, we can see that computational time increases very quickly when the training data are large-scale, hardly overlapping and non-linearly separable. In this case the most part of computations is wasted on the calculation of LM accuracy over all the training examples that significantly decreases the chance to find out the best LM for acceptable time. Below we describe our technique that can effectively learn multi-class concepts from large-scale EEG data.

## 4. A Neural-Network Decision Tree

The idea behind our method is to train decision tree nodes separately to classify all the pairs of classes. It follows that for the $r$ given classes the DT consists of $r*(r-1)/2$ TLUs which learn to classify the examples of two classes.

As we can see there are $(r-1)$ TLUs which deal with each of $r$ classes. These TLUs can be collected into a neural network so that the number of the networks becomes equal to $r$. The TLUs belonging to one neural network linearly approximate the dividing hyperplanes between the corresponding classes.

Formally, we can define a TLU, $f_{i/j}$, as a linear test which learns to divide the training examples between $i$th and $j$th classes, $\Omega_i$ and $\Omega_j$. The output of the TLU, $y$, is described as follows:

$$y = f_{i/j}(\pmb{x}) = \phantom{-}1, \ \forall \ \pmb{x} \in \Omega_i,$$
$$y = f_{i/j}(\pmb{x}) = -1, \ \forall \ \pmb{x} \in \Omega_j,$$

Let us now assume a case of $r = 3$ overlapping classes $\Omega_1$, $\Omega_2$, and $\Omega_3$ with centers $C_1$, $C_2$ and $C_3$ as depicted in Fig. 4. The number of TLUs as well as the neural networks, therefore, is equal to 3. In Fig. 4 the lines $f_{1/2}$, $f_{1/3}$ and $f_{2/3}$ depict the dividing hyperplanes that the TLUs perform. These hyperplanes divide the examples of classes $\Omega_1$ and $\Omega_2$, $\Omega_1$ and $\Omega_3$, $\Omega_2$ and $\Omega_3$, respectively. Note that a region of positive output values of TLUs is pointed here by an arrow.

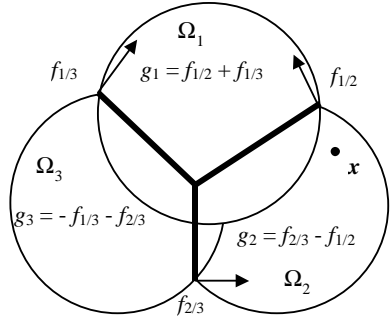

**Fig. 4:** Hyperplanes $g_1$, $g_2$, and $g_3$ dividing classes $\Omega_1$, $\Omega_2$, and $\Omega_3$

Grouping hyperplanes $f_{1/2}$, $f_{1/3}$ and $f_{2/3}$ together, we can build up the new dividing hyperplanes $g_1$, $g_2$, and $g_3$ as depicted in Fig. 4. The first hyperplane $g_1$ is a superposition of the linear tests $f_{1/2}$ and $f_{1/3}$: $g_1 = f_{1/2} + f_{1/3}$. These tests are taken with weights equal to 1 because $f_{1/2}$ and $f_{1/3}$ give the positive output values on the examples of class $\Omega_1$. Correspondingly, the second and third hyperplanes are $g_2 = f_{2/3} - f_{1/2}$ and $g_3 = -f_{1/3} - f_{2/3}$.

Assuming that an example **x** belongs to the class $\Omega_2$, we can see that **x** causes the outputs of $g_1$, $g_2$, and $g_3$ equal to 0, 2 and –2, respectively:

$$g_1(\mathbf{x}) = f_{1/2}(\mathbf{x}) + f_{1/3}(\mathbf{x}) = 1 - 1 = 0,$$
$$g_2(\mathbf{x}) = f_{2/3}(\mathbf{x}) - f_{1/2}(\mathbf{x}) = 1 + 1 = 2,$$
$$g_3(\mathbf{x}) = -f_{1/3}(\mathbf{x}) - f_{2/3}(\mathbf{x}) = -1 - 1 = -2.$$

As the output of $g_2(\mathbf{x})$ is larger than the others, the DT using the WTA strategy assigns correctly the given example **x** to the class $\Omega_2$.

To superpose the dividing hyperplanes $g_1, \ldots, g_r$, we can use the two-layer feed-forward neural networks consisting of $(r - 1)$ hidden TLUs, which are connected to the relevant input variables. Summing the contributions of these hidden neurons the output neurons make a final decision.

For an example discussed above, the DT depicted in Fig 5 consists of three neural networks and hidden neurons performing the linear tests $f_{1/2}$, $f_{1/3}$, and $f_{2/3}$ with weights $\mathbf{w}^{(1)}$, $\mathbf{w}^{(2)}$, and $\mathbf{w}^{(3)}$. As it follows from this example, the output neurons $g_1$, $g_2$, and $g_3$ are connected to the hidden neurons with weights equal to (+1, +1), (–1, +1) and (–1, –1), respectively.

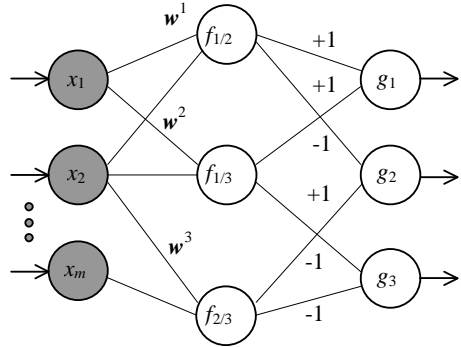

**Fig 5:** An example of the neural network decision tree for a case of $r = 3$ classes

In general case for $r > 2$ classes, DT consists of $r(r-1)/2$ hidden neurons $f_{1/2}, \ldots, f_{i/j}, \ldots, f_{r-1/r}$ and $r$ output neurons $g_1, \ldots, g_r$, where $i < j = 2, \ldots, r$. So, the weights of output neuron $g_i$ connected to the hidden neurons $f_{i/k}$ and $f_{k/i}$ are equal to + 1 and –1, respectively. We applied this neural network technique to our EEG problem and below discuss the result.

## 5. Experiments and Results

To learn the 16-class concept from the clinical EEGs, we applied the neural-network technique described above. For training and testing the concept, we used 39399 and 19670 EEG segments, respectively. For the given $r = 16$ classes, the DT consists of $r(r-1)/2 = 120$ linear tests.

The induced DT concept correctly classified the 80.8% of the training and 80.1% of the testing examples. Summing all the EEG segments belonging to one patient record, this concept correctly classified 89.2% and 87.7% of the 65 EEG records on the training and testing examples, respectively.

In Fig. 6, we plot the results of classifying the EEG testing segments over all 65 records. The number of the testing segments in these records varied between 30 and 780. Portions of segments which were correctly classified are depicted as the dark parts of bars. And portions of the incorrectly classified segments are depicted as the gray parts here. From this plot we can see that 8 records such as 4, 11, 29, 30, 35, 49, 52 and 65 were misclassified.

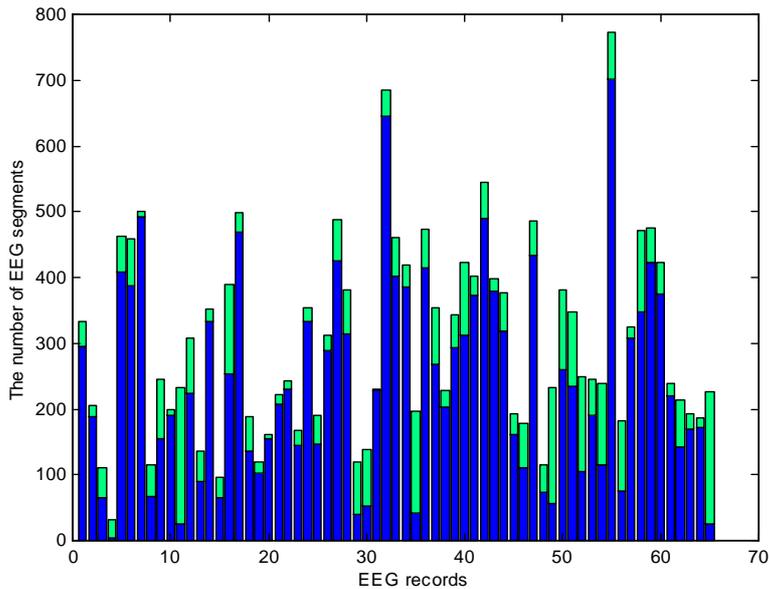

**Fig 6:** Results of testing the EEG segments over 65 records

In Fig. 7, we depict the summed outputs of the trained DT calculated for two patients over all the testing segments. As we can see, the summed outputs may interpret as the distributions of the EEG segments over all 16 classes. In this case, we can provide a probabilistic interpretation of making decisions. For example, we assign these patients to the second and third classes with probabilities 0.92 and 0.58, calculated as part of the correctly classified segments.

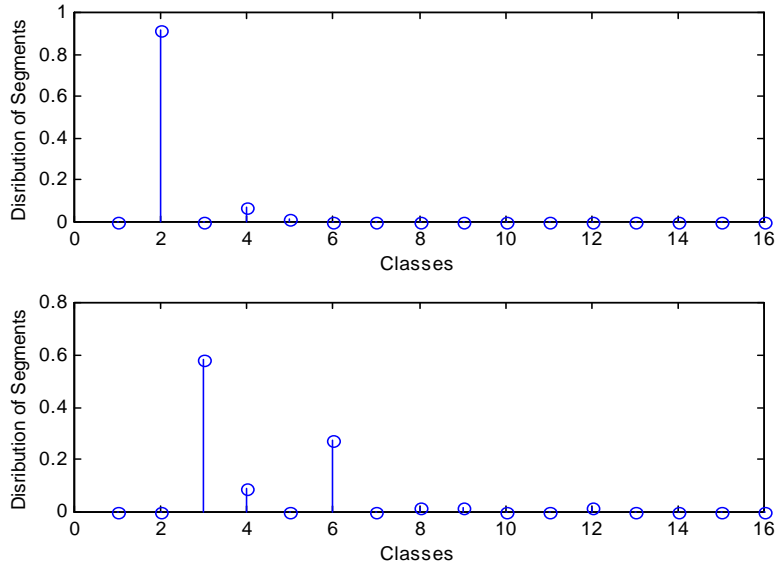

**Fig 7:** The summed outputs of the DT model for two newborns

## 6. Conclusion

The machine learning methods such as the LMs can not learn multi-class concepts well from large-scale EEG data. To learn multi-class concepts more successfully we developed a new technique based on the pairwise classification. The neural-network DT that we developed consists of TLUs which perform linear multivariate tests. The learning algorithm trains these tests separately, and then the output neurons linearly approximate the desired dividing hyperplanes.

This neural-network DT technique was applied to induce a 16-class concept from the clinical EEG data recorded from 65 newborns belonging to 16 age groups. The goal of this concept is to distinguish the brain maturation of the newborns aged between 31 and 51 weeks. Each EEG segment in these data was represented by 72 features. For training and testing the concept we used 39399 and 19670 EEG segments, respectively. As a result, the trained concept correctly classified 80.8% of the training and 80.1% of the testing examples, whilst the LM has correctly classified 71.6% of the testing data.

Thus we can conclude that our neural network DT technique is able to learn more accurate multi-class concepts from clinical EEG data. We believe also that this technique may be applied to other large-scale data.

**Acknowledgments**

The work has been supported by the University of Jena (Germany) during 2000-2002. The authors are grateful to Frank Pasemann for enlightening discussions, Joachim Frenzel and Burkhart Scheidt from the Pediatric Clinic of the University of Jena for their EEG records as well as to Jonathan Fieldsend from the University of Exeter (UK) for useful comments.